\documentclass{article}
\usepackage{spconf,amsmath,graphicx}
\usepackage{subcaption}
\usepackage{multirow}
\usepackage{tikz}

\title{A Spatially Constrained Deep Convolutional Neural Network for Nerve Fiber Segmentation in Corneal Confocal Microscopic Images using Inaccurate Annotations}
\name{Ning Zhang$^1$, Susan Francis$^2$, Rayaz A. Malik$^3$, Xin Chen$^4$}
\address{$^1$ Department of Electrical and Computer Engineering, University of British Columbia, Canada\\
         $^2$ Sir Peter Mansfield Imaging Centre, University of Nottingham, UK\\
         $^3$ Weill Cornell Medicine-Qatar, Qatar  \\
         $^4$ IMA Group, School of Computer Science, University of Nottingham, UK
         }
\begin{document}
%
\maketitle
\begin{abstract}
Semantic image segmentation is one of the most important tasks in medical image analysis. Most state-of-the-art deep learning methods require a large number of accurately annotated examples for model training. However, accurate annotation is difficult to obtain especially in medical applications. In this paper, we propose a spatially constrained deep convolutional neural network (DCNN) to achieve smooth and robust image segmentation using inaccurately annotated labels for training. In our proposed method, image segmentation is formulated as a graph optimization problem that is solved by a DCNN model learning process. The cost function to be optimized consists of a unary term that is calculated by cross entropy measurement and a pairwise term that is based on enforcing a local label consistency. The proposed method has been evaluated based on corneal confocal microscopic (CCM) images for nerve fiber segmentation, where accurate annotations are extremely difficult to be obtained. Based on both the quantitative result of a synthetic dataset and qualitative assessment of a real dataset, the proposed method has achieved superior performance in producing high quality segmentation results even with inaccurate labels for training.
\end{abstract}
\section{Introduction}
\label{sec:intro}

	Quantitative measurement of target objects (e.g. tumor, organ, \textit{etc.}) is very important for disease diagnosis in many medical applications. From classical image segmentation methods to more robust methods, various techniques (e.g. level-set \cite{chan2001active}, graph-cut \cite{boykov1999fast}, \textit{etc.}) have been proposed to achieve automatic image segmentation in a wide range of clinical problems. 

	Recently, deep convolutional neural network (DCNN) has achieved great success in medical image segmentation. Long \textit{et al.} have proposed the first end-to-end DCNN model, known as fully convolutional neural network (FCN) \cite{long2015fully}. Adapted from the FCN model, researchers have developed more and more advanced network structures, such as U-net \cite{ronneberger2015u} and SegNet \cite{badrinarayanan2017segnet} to achieve more accurate image segmentation. However, most of these models are based on fully supervised learning, where the performance is highly dependent on the quality of image annotations for training. These models do not have mechanisms to enforce smoothness and regional connectivity of segmented objects, which lead to discontinuously segmented objects. 
	
	Researchers have introduced the idea of traditional algorithms like Markov random field (MRF) and conditional random field (CRF) methods to DCNN models. Liu \textit{et al.} \cite{liu2015semantic} proposed a Deep Parsing Network to optimize MRF by modeling unary term and pairwise term using mean field theory. By solving MRF with a single feedforward pass, this efficient algorithm takes advantage of parallelized computing and pairwise information from neighbors. Chen et al. \cite{chen2014semantic} combined the last layer of DCNN model with CRF, but they performed as two separate components. Similarly, Kamnitsas \textit{et al.} \cite{kamnitsas2017efficient} applied CRF as a post-processing step to refine the prediction from DCNN by improving the smoothness of the segmentation output. Integrating CRF as recurrent neural network (RNN) into DCNN models enables joint training of both DCNN and CRF parameters. Zheng \textit{et al} \cite{zheng2015conditional} proposed a ``CRF-as-RNN" model which firstly train a DCNN model followed by continuous training of the CRF-as-RNN layers.
	
	These models successfully integrated probabilistic graphical models with deep learning methods which produced excellent smooth semantic segmentation results. However, to achieve the desired outcome, this refinement step requires either a good initial segmentation output produced by a DCNN model or an initially-trained DCNN model. Balancing the unary term and pairwise term of a graphical model in a separate procedure is not a trivial task. In most medical applications, high-quality annotations are extremely difficult and expensive to provide for model training. Therefore, the segmentation results from the first DCNN model may not be good enough for CRF-based second-stage training or refinement. None of the above studies discussed the effects of training the DCNN-CRF models based on inaccurate labels. 

	In this paper, we propose to add a spatially consistent term to the cost function (e.g. cross entropy) of conventional DCNN based segmentation models (e.g. U-net \cite{ronneberger2015u}). We integrate CRF as a dynamic convolutional layer into a DCNN model, which enables both the unary term and pairwise term of a graphical model to be trained simultaneously without any pre-training or post-processing steps. This prevents the model to be over-fitted to the training labels and improves the segmentation performance significantly when the annotated labels are not accurate. The proposed method has been evaluated based on confocal microscopic images (CCM) for nerve fiber segmentation, where only single-pixel skeletons for nerve fibers are manually annotated by experts and more accurate annotations are extremely difficult to be obtained. For quantitative evaluation, a set of synthetic CCM dataset with true ground truth annotations is generated. The main contribution of this paper is the integration of a spatially consistent constraint to DCNN models for overcoming the issue of inaccurate annotations for supervised learning.

\section{Method}
\label{sec:method}
\subsection{U-net}
Our proposed spatially constrained DCNN method is implemented based on a widely used network (i.e. U-net \cite{ronneberger2015u}), which consists of an encoding path and a decoding path. At each layer of the encoding path, two 3$\times$3 convolutions followed by a rectified linear unit (ReLU) are applied. A 2$\times$2 max pooling operation with stride 2 for down-sampling is then used after each layer which halves the feature map while doubling the number of feature channels. In the decoding path, we construct 2$\times$2 up-convolutional layers which up-sample the feature maps and half the number of feature channels. After each up-convolutional layer, a feature map from the corresponding layer in the encoding path is copied and concatenated to the up-sampled feature map. At the end of the decoding path, a 1$\times$1 convolutional layer is used to map the feature vector to the desired class of labels and produces the prediction. Based on the prediction, we also calculate the spatial consistent cost term with details described in the next section. Based on a weighted sum of the cross entropy cost and the spatially consistent cost, the error is backpropagated for network parameter optimization. Detailed parameter settings are reported in section \ref{sec:para}

\subsection{Spatially consistent cost}
In this section, we introduce how the spatially consistent cost term is derived from the idea of CRF. CRF is defined on observations $\mathbf{I}=\{I_1,\dots,I_N\}$ and the domain of the random field $\mathbf{X}=\{X_1,\dots,X_N\}$ with a set of latent variables $\mathbf{V}=\{V_1,V_2,\dots,V_K\}$. In the context of image segmentation, the CRF model can be constructed as: $\mathbf{I}$ ranges over image with $N$ pixels, $\mathbf{X}$ ranges over possible pixel-wise labeling and $\mathbf{L}$ represents a set of $K$ class labels. $I_{n}$ is the color (RGB images) or intensity (gray images) values of pixel $n$ and $X_{n}$ is the label assigned to pixel $n$. The problem of image segmentation can be solved as graphical model optimisation by minimising the energy function in equation (1).
\begin{eqnarray}
E(\mathbf{X})=\sum_{i} \phi_{u}\left(X_{i}| \mathbf{I}\right)+\sum_{i \neq j} \phi_{p}\left(X_{i}, X_{j} | \mathbf{I}\right)
\end{eqnarray}
$\phi_{u}$ and $\phi_{p}$ are the unary and pairwise terms respectively. For a binary class segmentation problem, $V=\{0,1\}$ that are associated with background and foreground respectively. A configuration $\mathbf{X}$ is one of the possible assignments of all $N$ pixels and the ground truth labels (denoted as $\mathbf{Y}$) is one of them. The unary term (first term in equation (1)) is computed independently for each pixel by a classifier, given an input image. We use U-net based classification model (section 2.1), and the cost function in this case is the cross entropy cost. The pairwise potential $\phi_p$ represents the penalty of assigning labels to the current pixel $i$ and another pixel $j$ in the graph at the same time. Here, we propose to use a local CRF model, where the pairwise term is determined by its eight nearest neighbor pixels. This enables a local smoothness constraint to the segmentation results. The pairwise term (second term in equation (1)) of the CRF model is defined as below.
\begin{eqnarray}
\psi_{i, j}=\mu_{i, j} P_{i} P_{j}
\end{eqnarray}
\begin{eqnarray}
\mu_{i,j}=\left\{
\begin{array}{rcl}
&-\exp(-\frac{(I_{i}-I_{j})^{2}}{2 \sigma^{2}})       &(j \in R, \hat{Y}_{i}=\hat{Y}_{j})   \\
&+\exp (-\frac{(I_{i}-I_{j})^{2}}{2 \sigma^{2}})     &(j \in R, \widehat{Y}_{i} \neq \widehat{Y}_{j})   \\
&0     &(j \notin R)   
\end{array} \right.
\end{eqnarray}
where $P_i$ and $P_j$ are the predicted confidence scores for central pixel $i$ and its neighbor pixel $j$ respectively from the U-net model at the current training iteration. $\mu_{i,j}$ is the weight between them. $\hat{Y}_{i}$ and $\hat{Y}_{j}$ are the predicated labels from U-net model at current training iteration for pixel $i$ and $j$ respectively. $R$ denotes the domain of the eight nearest neighbors for central pixel $i$. For a single pixel, we calculate the spatially consistent cost as the normalized marginal probability using equation (4).
\begin{eqnarray}
\psi_{i}=\frac{\sum_{j} \psi_{i, j}}{\sum_{j} |\mu_{i, j}|}
\end{eqnarray}
Overall, the total cost function that combines both the unary term (cross entropy) and pairwise term is:
\begin{eqnarray}
L=\frac{1}{N} \sum_{i}-\log P\left(X_{i}=Y_{i} | \mathbf{I}\right)+\lambda \sum_{i}\psi_{i}
\end{eqnarray}
$\lambda$ is a parameter for balancing the two terms. $L$ in equation (5) is minimised by using the U-net model described in section 2.1. Detailed implementation is available on GitHub \footnote{https://github.com/XinChenNottingham/SpatiallyConstrainedDCNN}.

\section{Experiments and Results}
We evaluated the performance of our method on a binary image segmentation task using CCM images for nerve fiber segmentation. Only single-pixel skeletons for nerve fibers were manually annotated by experts for model training. An accurate fiber thickness was extremely difficult to obtain. This inaccurate annotation poses a particular challenging problem compared to other public datasets where accurate annotations are provided. For quantitative evaluation, a set of synthetic CCM dataset with `true’ ground truth annotation is generated. Qualitative results of real CCM images are also provided.

\subsection{Dataset}
\subsubsection{CCM dataset}
\begin{figure}
	\begin{center}
		\begin{subfigure}[normal]{0.15\textwidth}
			\includegraphics[width = \textwidth]{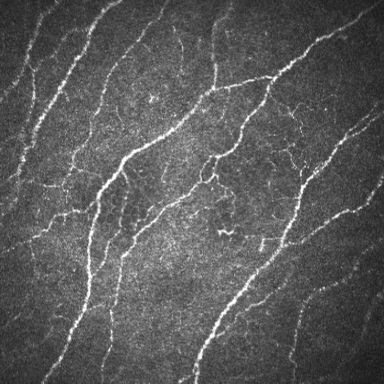}
			\caption{}
			\label{fig:ccm:org}
		\end{subfigure}
		\vspace{0.1 in}
		\begin{subfigure}[normal]{0.15\textwidth}
			\includegraphics[width = \textwidth]{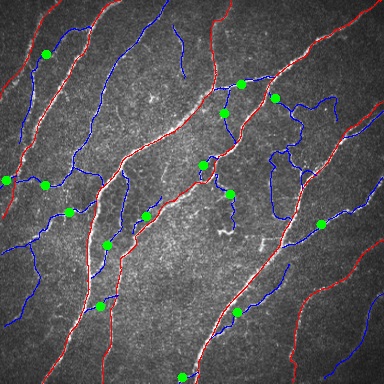}
			\caption{}
			\label{fig:ccm:ano}
		\end{subfigure}
		\vspace{0.1 in}
		\begin{subfigure}[normal]{0.15\textwidth}
			\includegraphics[width = \textwidth]{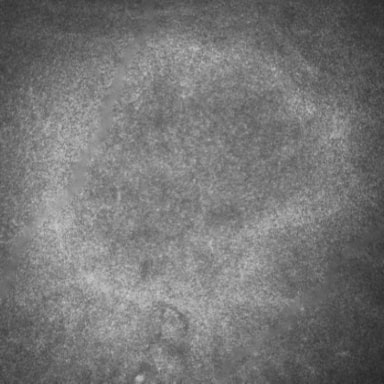}
			\caption{}
			\label{fig:syn:a}
		\end{subfigure}
		\vspace{-15pt}
		\vspace{0.1 in}
		\begin{subfigure}[normal]{0.15\textwidth}
			\includegraphics[width = \textwidth]{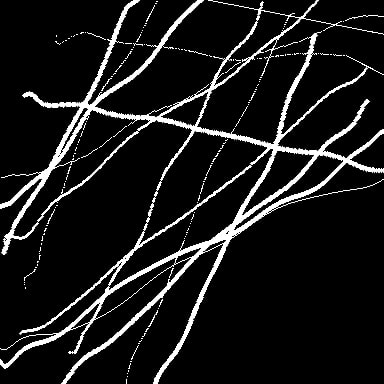}
			\caption{}
			\label{fig:syn:b}
		\end{subfigure}
		\vspace{0.1 in}
		\begin{subfigure}[normal]{0.15\textwidth}
			\includegraphics[width = \textwidth]{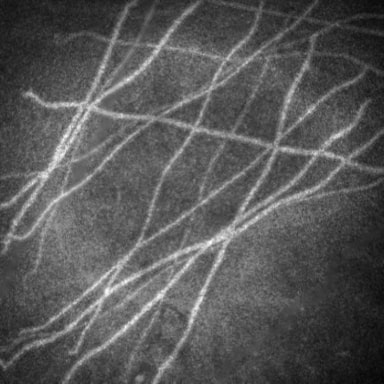}
			\caption{}
			\label{fig:syn:c}
		\end{subfigure}
	    \vspace{0.1 in}
		\begin{subfigure}[normal]{0.15\textwidth}
			\includegraphics[width = \textwidth]{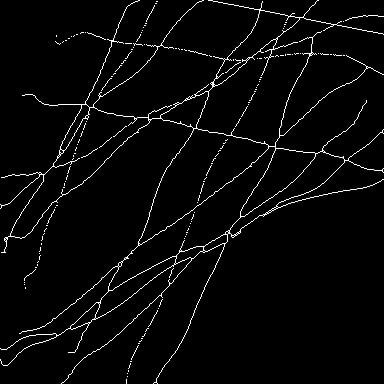}
			\caption{}
			\label{fig:syn:d}
		\end{subfigure}
	\end{center}
	\vspace{-25pt}
	\caption{(a) CCM image. (b) Manual annotation. Both red lines and blue lines are considered as nerve fibers for our segmentation task. (c) Background image extracted from a real CCM image. (d) The generated polynomial curves with random widths, lengths and orientations. (e) A synthetic CCM image. (f) Single pixel pseudo label for model training.}
	\label{fig:syn}
	\vspace{-10pt}
\end{figure}

Corneal confocal microscopy images of nerve fibers were captured from the subbasal plexus immediately above Bowman’s membrane of the cornea by an in-vivo laser confocal microscope. Clinical studies have shown that CCM is capable of making quantitative assessment of various neurodegenerative diseases (e.g. diabetic neuropathy \cite{chenDB2015}). In this study, CCM images (Fig.\ref{fig:ccm:org}) were captured from 176 subjects (84 healthy and 92 patients with Type 1 diabetes) using the Heidelberg Retina Tomograph Rostock Cornea Module (HRT-III). The image dimensions are $384\times384$ pixels with the pixel size of $1.0417\mu m$. All nerve fibers in the CCM images were manually traced by an experienced clinician using single-pixel lines (Fig.\ref{fig:ccm:ano}). A total of 949 CCM images (4-6 images per subject) are available for our experiments.
\vspace{-10pt}

\subsubsection{Synthetic dataset}
 To generate realistic CCM images for quantitative evaluation, the following steps are followed. (1) Apply the image in-painting method described in \cite{bertalmio2000image} to remove foreground pixels from real CCM images (e.g. Fig.\ref{fig:ccm:org}), resulting in a background CCM image (Fig.\ref{fig:syn:a}). (2) Randomly create a number of polynomial curves with different widths, lengths and orientations as the synthetic nerve fibers (Fig.\ref{fig:syn:b}). (3) Smooth the synthetic nerve fibers with a Gaussian filter and add them to the background CCM image in Fig.\ref{fig:syn:a}, resulting in a completed synthetic CCM image as shown in Fig.\ref{fig:syn:c}. Furthermore, the true nerve fiber annotation in Fig.\ref{fig:syn:b} is eroded to single-pixel skeletons and randomly shifted by $\pm3$ pixels to mimic the human annotations for model training, as shown in Fig.\ref{fig:syn:d}. We have generated 1000 synthetic images for method evaluation and comparison, and an extra 500 validation images for parameter optimisation. 

\subsection{Parameter settings}
\label{sec:para}
The proposed network architecture is the same as described in section 2.1. We used 5 layers in both encoding and decoding paths. Batch normalization \cite{ioffe2015batch} and dropout \cite{srivastava2014dropout} with rate of $25\%$ on all convolutional layers were applied. We trained our network using Adam optimizer with an initial learning rate of $10^{-4}$. All networks were trained for 300 epochs where the models were converged. We have experimentally tested different values for $\sigma$ in equation (3) and $\lambda$ in equation (5) using the 500 independent validation images. Based on our experiments, $\sigma=0.5$ and $\lambda = 1$ achieved the best performance and consistently used for all experiments in our evaluations. We implemented our framework using the TensorFlow library. All input images were resized to $512\times512$. The models were trained using a single Nvidia GTX 1080Ti graphic card.
\subsection{Evaluation Results}
\subsubsection{Synthetic dataset}
Based on the 1000 synthetic CCM images, we randomly selected 500 for training and 500 for testing. We compared our method with U-net, U-net with fully-connected CRF \cite{krahenbuhl2011efficient} as post-processing (U-net+CRF) and the current state-of-the-art method for CCM nerve fiber segmentation using a classical machine learning method (Chen \cite{chen2016automatic}). Chen's method is based on handcrafted features and random forest classifier. Dice coefficient (DC), precision and recall, calculated by comparing the binary predictions of these methods with the ground truth images, are listed in Table \ref{tabel:1}. Values for ``Baseline" in Table \ref{tabel:1} are calculated using the pseudo labels for model training (Fig. \ref{fig:syn:d}) and the `true' ground truth labels (Fig. \ref{fig:syn:b}). 

\begin{table} [h]
	\caption{Quantitative evaluation results for the synthetic dataset. Values of mean $\pm$ standard deviation are reported.}
	\vspace{-15pt}
	\begin{center}
		\begin{tabular}{ |c|c|c|c|c| } 
			\hline
			Method&DC&Precision &Recall\\
			\hline
			Baseline & $0.38\pm 0.04$ & $0.94\pm 0.06$  & $0.24\pm 0.03$\\
			Chen \cite{chen2016automatic} & $0.67\pm 0.12$ & $0.58\pm 0.15$  & $0.84\pm 0.07$\\
			U-net & $0.60\pm 0.10$& $0.98\pm 0.02$& $0.45\pm 0.03$\\
			U-net+CRF & $ 0.64\pm0.13 $& $ 0.97\pm0.03 $& $0.50\pm0.04$\\
			Proposed& $0.80 \pm 0.12$& $0.90\pm 0.04$ & $ 0.75\pm0.06 $\\
			\hline
		\end{tabular}
	\end{center}
	\label{tabel:1}
	\vspace{-20pt}
\end{table}

From Table.\ref{tabel:1}, it can be seen that the DC and recall values of U-net are lower than Chen's method but with a much higher precision value. Chen's method is based on handcrafted features which is more generic to detect linear structures but resulting in more false positives than U-net hence lower precision. The high precision value of U-net is due to that it learns to detect single pixel labels that are well within the nerve fiber region. However, it performs poorly in recovering the nerve fibre width. The U-net+CRF method refines the U-net outputs slightly that improves the performance in terms of DC and recall measurements with similar precision values. We observed that CRF post-processing successfully joint disconnected nerve fibers but failed to expand the labels to their neighbouring regions. This was due to the converged U-net model had very high confidence values for the detected background pixels even in the regions surrounding the predicted single pixel label locations. Hence it was difficult for the pair-wise term in CRF to change the labels. An early stop of U-net training may overcome this issue, but robustly determining the stopping point and parameter setting for CRF is not trivial. The proposed method has achieved significantly (statistical significance using Wilcoxon signed-rank test with $p<0.001$) better performance than the U-net, U-net+CRF and Chen's method based on DC measurements. Our method achieves overall the best performance which prevents the model being over-fitted to the training data by connecting neighbouring pixels, which results in more false positives ($7\%-8\%$ lower precision) but much less false negatives ($25\%-30\%$ higher recall) than U-net and U-net+CRF. 

\subsubsection{Real dataset}
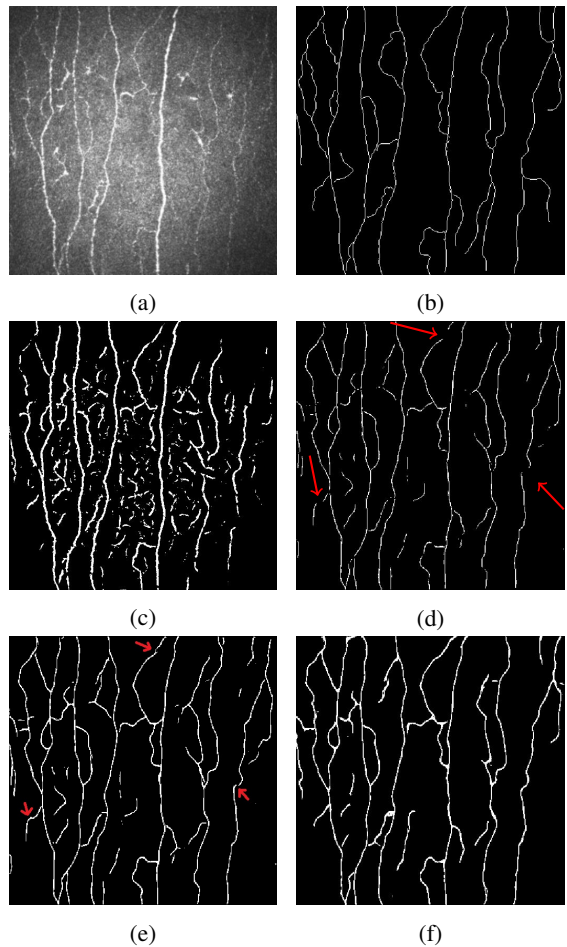
\begin{figure}[h]
	\begin{center}
		\begin{subfigure}[normal]{0.2\textwidth}
			\includegraphics[width = \textwidth]{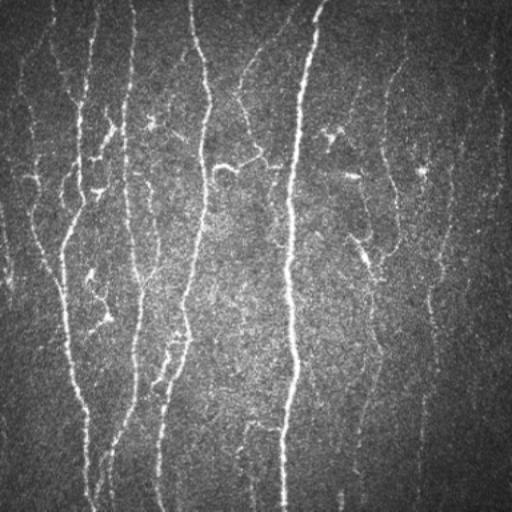}
			\caption{}
			\label{fig:assCCM:a}
		\end{subfigure}
		~
		\begin{subfigure}[normal]{0.2\textwidth}
			\includegraphics[width = \textwidth]{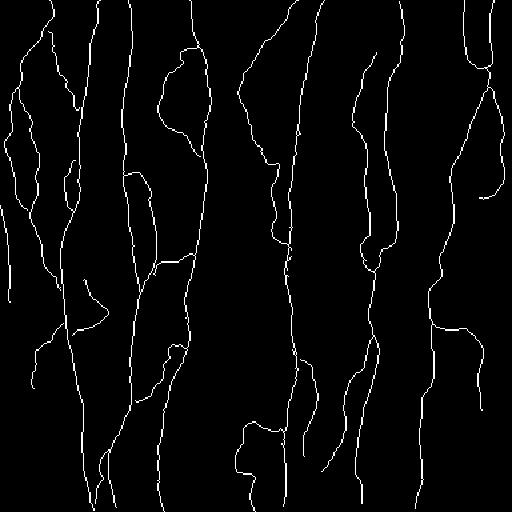}
			\caption{}
			\label{fig:assCCM:b}
		\end{subfigure}
		~
	    \begin{subfigure}[normal]{0.2\textwidth}
			\includegraphics[width = \textwidth]{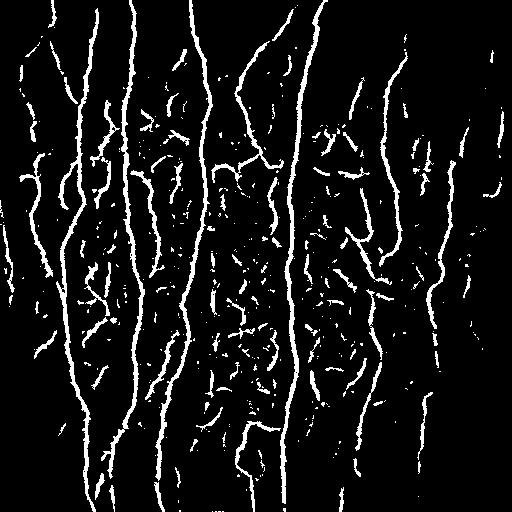}
			\caption{}
			\label{fig:assCCM:c}
		\end{subfigure}
		~
		\begin{subfigure}[normal]{0.2\textwidth}
			\begin{tikzpicture}[remember picture] 
			\node[anchor=south west,inner sep=0] (imageA) {\includegraphics[height=\textwidth]{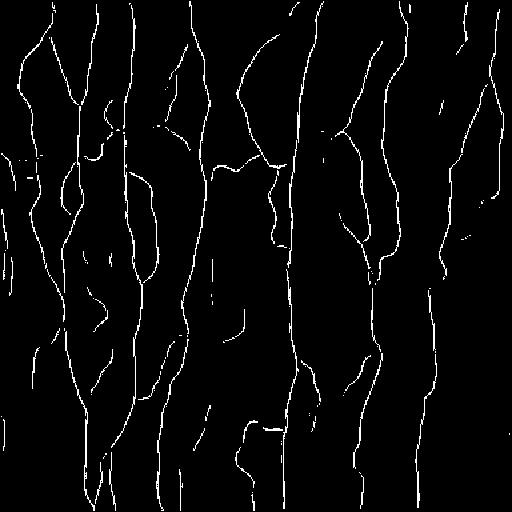}}; 
			\begin{scope}[x={(imageA.south)},y={(imageA.west)}]
			\node[coordinate] (B) at (0,0) {}; 
			\draw[color=red,thick,->] (0.1,1) -- (0.16,0.7);
			\draw[color=red,thick,->] (1.99,0.6) -- (1.8,0.8);
			\draw[color=red,thick,->] (0.7,1.99) -- (1.05,1.9);
			\end{scope} 
			\end{tikzpicture}
			\caption{}
			\label{fig:assCCM:d}
		\end{subfigure}
		~ \begin{subfigure}[normal]{0.2\textwidth}
			\includegraphics[width = \textwidth]{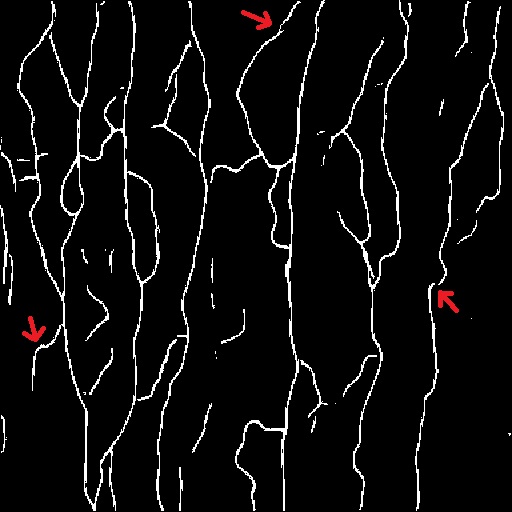}
			\caption{}
			\label{fig:assCCM:e}
		\end{subfigure}
		~ \begin{subfigure}[normal]{0.2\textwidth}
			\includegraphics[width = \textwidth]{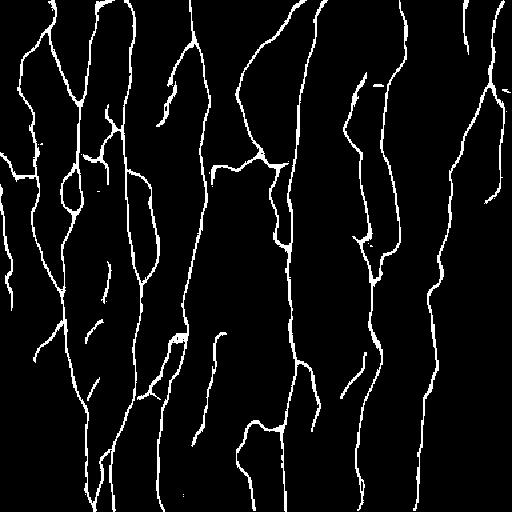}
			\caption{}
			\label{fig:assCCM:f}
		\end{subfigure}
	\end{center}
	\vspace{-10pt}
	\caption{Qualitative results of an example of real CCM image. (a) Original CCM image. (b) Inaccurate human annotation for training. (c) Result of Chen's method. (d) Result of U-net (e) Result of U-net+CRF (f) Result of our method.}
	\label{fig:assCCM}
	\vspace{-10pt}
\end{figure}

For the real CCM dataset, we used 500 images for training and 449 images for testing. One randomly selected example CCM image and its manual annotation are shown in Fig.\ref{fig:assCCM:a} and Fig.\ref{fig:assCCM:b} respectively. Similar to the conclusion drawn from the quantitative results, Chen's method (Fig.\ref{fig:assCCM:c}) detects more false positives than other methods. U-net method predicts very thin nerve fibers that are similar to the training labels, but with some discontinuities that are highlighted by red arrows in Fig.\ref{fig:assCCM:d}. The U-net+CRF method provides better visual results than U-net by connecting some of the disconnected nerve fibres (indicated by red arrows in Fig.\ref{fig:assCCM:e}). Obviously, our method (Fig.\ref{fig:assCCM:f}) generates smoother and more accurate segmentation results in recovering nerve fiber width. It is also worth noting that the U-net+CRF method is quite sensitive to parameter settings (refer to \cite{krahenbuhl2011efficient}), which is not trivial to find the best parameters based on visual preference. In contrast, the parameter $\sigma$ and $\lambda$ of our method are fixed throughout all experiments and no post-processing needed. 

\vspace{-10pt}
\section{Discussion and Conclusions}
In conclusion, we present an effective image segmentation method which integrates local spatial constrains to the U-net DCNN model. This prevents the DCNN model being over-fitted to the training data when the training labels are not accurately annotated. Based on a challenging CCM dataset, we have shown that our method significantly outperforms the conventional U-net, U-net with CRF post-processing and the current state-of-the-art method. For future work, we will extend this method to multi-class image segmentation and 3D images for different medical applications. 

\label{sec:ref}
\bibliographystyle{IEEEbib}
\bibliography{refs}
\end{document}